\pdfoutput=1
\documentclass[11pt]{article}
\usepackage[preprint]{acl}

\usepackage{times}
\usepackage{latexsym}
\usepackage[T1]{fontenc}
\usepackage[utf8]{inputenc}
\usepackage{microtype} %
\usepackage{inconsolata}
\usepackage{graphicx}

\usepackage{amsmath}
\usepackage{amssymb}
\usepackage{booktabs}
\usepackage{enumitem}
\usepackage{tikz}
\usetikzlibrary{shapes, shapes.geometric, arrows.meta, positioning}

\newcommand{\tabstyle}{\small\setlength{\tabcolsep}{2pt}}

\hypersetup{hypertexnames=false}

\makeatletter
\newif\ifanon
\ifacl@anonymize\anontrue\else\anonfalse\fi
\makeatother
\ifanon\pdfinfoomitdate=1 \pdfsuppressptexinfo=-1 \fi

\newif\ifpending
\pendingfalse

\newcommand{\prov}[1]{#1}

\usepackage{soul}
\sethlcolor{yellow}
\definecolor{hlpink}{RGB}{255,182,193}
\newif\ifhlmark\hlmarkfalse
\ifhlmark\else\fi

\usepackage{colortbl}
\definecolor{groupblue}{RGB}{221,235,247}

\newcommand{\lgo}[1]{\raisebox{-0.55pt}{\includegraphics[height=6pt]{plots/logos/#1.png}}}

\newcommand{\best}[1]{\textbf{\boldmath #1}}
\newcommand{\btext}[1]{\underline{#1}}
\newcommand{\bestkey}[1]{\textbf{#1}}
\newcommand{\btextkey}[1]{\underline{#1}}

\newcommand{\ccFemB}{\prov{0.991}} 
\newcommand{\ccAbiB}{\prov{0.864}} 
\newcommand{\ccIntB}{\prov{0.682}} 
\newcommand{\ccExtB}{\prov{0.699}} 
\newcommand{\ccFemM}{\prov{0.991}} 
\newcommand{\ccAbiM}{\prov{0.864}} 
\newcommand{\ccIntM}{\prov{0.693}} 
\newcommand{\ccExtM}{\prov{0.693}} 
\newcommand{\ccFemT}{\prov{0.985}} \newcommand{\ccFemC}{\prov{0.583}}
\newcommand{\ccAbiT}{\prov{0.827}} \newcommand{\ccAbiC}{\prov{0.723}}
\newcommand{\ccIntT}{\prov{0.658}} \newcommand{\ccIntC}{\prov{0.742}}
\newcommand{\ccExtT}{\prov{0.623}} \newcommand{\ccExtC}{\prov{0.769}}

\title{No Transformer Beats Six Covariates: Predicting Depressive Symptoms at Age 23 from Age-11 Essays}

\author{%
 \textbf{Daniel Kua\textsuperscript{1}},
 \textbf{Emrul Hasan\textsuperscript{2,3}},
 \textbf{John-Jose Nunez\textsuperscript{3}},
 \textbf{Frances Chen\textsuperscript{4}}
\\
 \textsuperscript{1}Department of Statistics, University of British Columbia, Vancouver, Canada \\
 \textsuperscript{2}Vector Institute, Toronto, Canada \\
 \textsuperscript{3}Department of Psychiatry, University of British Columbia, Vancouver, Canada \\
 \textsuperscript{4}Department of Psychology, University of British Columbia, Vancouver, Canada \\
 \small\texttt{daniel.kua@stat.ubc.ca}, \texttt{emrul.phy@gmail.com},
 \texttt{johnjose.nunez@ubc.ca}, \texttt{frances.chen@psych.ubc.ca}
}

\renewcommand{\topfraction}{0.92}

\renewcommand{\textfraction}{0.08}
\renewcommand{\floatpagefraction}{0.95}

\begin{document}
\maketitle

\begin{abstract}
Natural language processing (NLP) models can detect depression-related
language in text written near the time symptoms are measured, but whether
pretrained transformers can predict depressive symptoms from text written
twelve years earlier is largely untested. In the National Child Development
Study, a British birth cohort, we predict probable depressive symptoms at age
23 from essays the same people wrote at age 11. Our baseline, a logistic
regression on six childhood covariates, outperforms every text model that sees
only the essay: seven fine-tuned transformers, a bag-of-words model, frozen
embeddings and four zero-shot large language models. Its area under the
receiver operating characteristic curve (AUC-ROC) is 0.737, against 0.670 for
the best transformer on the primary seed, and no added text score detectably
raises the baseline's AUC-ROC. None of the five domain-pretrained transformers
detectably beats its general-domain control after Bonferroni correction. For
long-horizon prediction, the baseline remains the model to beat.
\end{abstract}

\section{Introduction}

\begin{figure*}[t]
\centering
\resizebox{\textwidth}{!}{%
\begin{tikzpicture}[
 font=\small,
 box/.style={rectangle, draw, rounded corners,
 minimum height=0.85cm, align=center, inner sep=4pt},
 arrow/.style={-{Stealth[length=1.6mm]}, semithick}
]
\node[box, fill=gray!15, minimum width=6.4cm] (essay) at (0,0)
 {Raw essay text (written at age 11)};
\node[box, fill=green!14, minimum width=5.0cm] (cov) at (8.6,0)
 {Six childhood covariates\\(no essay text)};
\node[box, fill=blue!12, minimum width=11.0cm] (enc) at (0,-1.55)
 {\textbf{Four ways to turn the essay into one score}\\[2pt]
 TF-IDF $\mid$ frozen BERT/BGE embeddings (each $+$ logistic regression)\\
 zero-shot ratings from 4 LLMs $\mid$ 7 fine-tuned transformers\\
 (general, domain-specific, cross-condition)};
\node[box, fill=orange!16, minimum width=5.0cm] (lr) at (8.6,-1.55)
 {Logistic regression};
\node[box, fill=orange!20, minimum width=6.4cm] (head) at (0,-3.35)
 {Text score for each essay};
\node[box, fill=green!20, minimum width=13.8cm] (out) at (4.3,-4.45)
 {\textbf{Predicted risk of depressive symptoms at age 23}};
\draw[arrow] (essay) -- (enc);
\draw[arrow] (enc) -- (head);
\draw[arrow] (head) -- (head |- out.north);
\draw[arrow] (cov) -- (lr);
\draw[arrow] (lr) -- (lr |- out.north);
\draw[arrow, dashed] (head.east) to[out=0, in=270] ([xshift=-10mm]lr.south);
\end{tikzpicture}%
}
\caption{Text models against the covariate baseline. The dashed path adds the text score to the covariates.}
\label{fig:architecture}
\end{figure*}
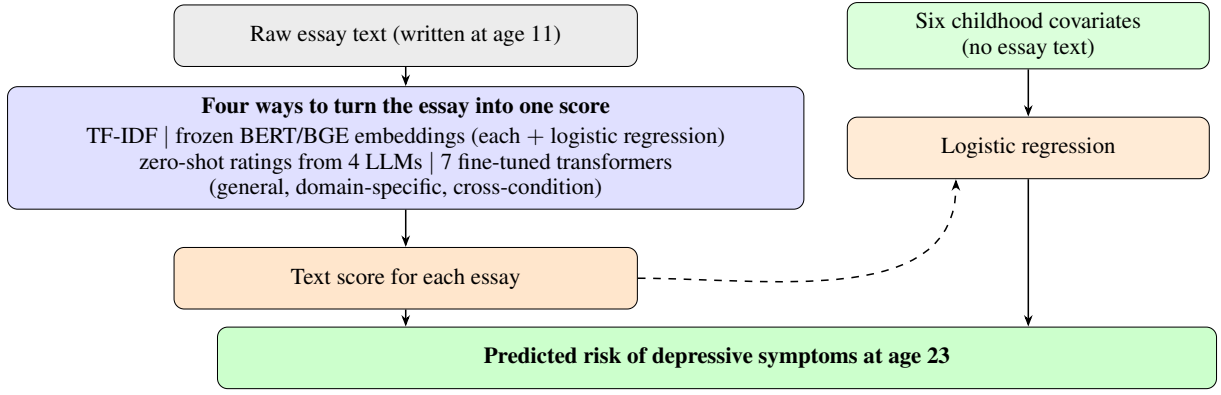

Many mental-health natural language processing (NLP) studies evaluate models on text
written close to the outcome \citep{coppersmith2015clpsych, de2013predicting, eichstaedt2018facebook}.
In that near-term setting, transformers such as BERT \citep{devlin2019bert}
and domain-pretrained variants such as MentalBERT \citep{ji2022mentalbert}
perform well, yet early intervention calls for a harder task: predicting
risk before symptoms appear.
Because depression often begins in adolescence
\citep{thapar2022depression}, a model that uses childhood data must bridge a
gap of years, across which pretrained transformers are largely untested.

We test them in the National Child Development Study (NCDS), a British birth
cohort of over 17{,}000 people followed since 1958 \citep{power2006cohort}.
At age 11, participants wrote essays imagining life at 25. Twelve years later,
at age 23, they completed the Malaise Inventory, a screen for probable
depressive symptoms on which 8 or more out of 24 counts as positive.

For long-horizon prediction, text is useful only if it adds to childhood
covariates that cohorts already collect.
\citet{boehm2024words} found that hand-coded well-being scores from these
essays were associated with later depressive symptoms after adjusting for
those covariates. Because hand-coding does not scale, we test text models that
see only the essay and score it (Figure~\ref{fig:architecture}): seven
fine-tuned transformers, term frequency-inverse document frequency (TF-IDF)
features, frozen embeddings and four zero-shot large language models (LLMs).
The covariate baseline is a logistic regression on six childhood covariates:
sex, father's social class, parental psychiatric history, cognitive ability
and two behaviour ratings. We draw on a reporting guideline for clinical
prediction models \citep{collins2024tripod} and ask three research questions
(RQs). RQ1 asks whether a text model alone can match six childhood covariates,
RQ2 whether any text score adds to them, and RQ3 whether domain-pretrained
transformers help. Our contributions are:
\begin{itemize}[leftmargin=1.2em,topsep=2pt,partopsep=0pt,itemsep=1pt,parsep=0pt]
\item \textbf{RQ1: The baseline beats every text model}. Its area under the
 receiver operating characteristic (ROC) curve (AUC-ROC) is $0.737$, against
 $0.670$ for the best transformer on the primary seed
 (Table~\ref{tab:results}). In paired tests, it leads all seven transformers
 and their seed ensembles, five of each after Bonferroni correction
 (Figure~\ref{fig:paired}).
\item \textbf{RQ2: No added text score detectably raises the baseline's
 AUC-ROC}, whether trained (cross-fitted) or zero-shot
 (Section~\ref{sec:incremental}, Table~\ref{tab:incremental}). Two
 transformers can read these essays, recovering sex and cognitive ability at
 age 11 (Table~\ref{tab:concurrent}), but the trained scores carry little that
 predicts the outcome beyond the six covariates (Table~\ref{tab:mechanism}).
\item \textbf{RQ3: No domain-pretrained transformer detectably beats its
 general-domain control} after Bonferroni correction
 (Section~\ref{sec:pairwise}, Figure~\ref{fig:paired}), and all five
 seed-ensemble intervals include zero. Task-adaptive pretraining on the
 essays slightly raises AUC-ROC, but no gain survives Bonferroni correction
 (Section~\ref{sec:closing}).
\end{itemize}

\begin{table}[t]
\centering
\tabstyle
\begin{tabular*}{\columnwidth}{@{\extracolsep{\fill}}lr@{}}
\toprule
Step & N \\
\midrule
Essays parsed, non-empty (UKDA-8313) & 10{,}509 \\
Plus sex available (UKDA-5565) & 10{,}507 \\
Plus all six childhood covariates & 9{,}333 \\
\textbf{Plus Malaise at 23 (UKDA-5566)} & \textbf{7{,}226} \\
\midrule
Below the cut (Malaise $< 8$) & 6{,}729 (93.1\%) \\
At or above the cut (Malaise $\geq 8$) & 497 (6.9\%) \\
\bottomrule
\end{tabular*}
\caption{Cohort flow. \textbf{Bold}: analytic sample.}
\label{tab:cohort_flow}
\end{table}
\begin{table*}[!t] %
\centering
\tabstyle
\begin{tabular*}{\textwidth}{@{\extracolsep{\fill}}lcccc@{}}
\toprule
Model & AUC-ROC\,$\uparrow$ [95\% CI] & AUC-ROC$_{\text{5-run}}$\,$\uparrow$ & AUC-PR\,$\uparrow$ & Brier\,$\downarrow$ \\
\midrule
Covariate baseline (logistic regression) & \best{0.737} [0.688, 0.783] & -- & \best{0.188} & \best{0.060} \\
TF-IDF + logistic regression (non-neural) & 0.644 [0.591, 0.697] & -- & 0.115 & 0.063 \\
\midrule
\rowcolor{groupblue}\multicolumn{5}{l}{General-domain language models} \\
BERT-base-uncased & 0.669 [0.617, 0.718] & 0.663\,$\pm$0.026 & 0.128 & 0.064 \\
RoBERTa-base & 0.620 [0.567, 0.670] & 0.661\,$\pm$0.005 & 0.097 & 0.063 \\
\rowcolor{groupblue}\multicolumn{5}{l}{Domain-specific language models} \\
ClinicalBERT & 0.667 [0.612, 0.718] & 0.637\,$\pm$0.016 & 0.144 & 0.063 \\
BioBERT & \btext{0.670} [0.616, 0.723] & 0.651\,$\pm$0.017 & 0.146 & 0.063 \\
MentalBERT & 0.619 [0.562, 0.677] & \best{0.671}\,$\pm$0.007 & 0.142 & 0.065 \\
MentalRoBERTa & 0.646 [0.591, 0.700] & 0.654\,$\pm$0.011 & \btext{0.149} & \btext{0.063} \\
\rowcolor{groupblue}\multicolumn{5}{l}{Cross-condition transfer \citep{amalq2025stress}} \\
StressRoBERTa & 0.660 [0.606, 0.712] & 0.650\,$\pm$0.006 & 0.135 & 0.066 \\
\rowcolor{groupblue}\multicolumn{5}{l}{Zero-shot LLMs (not fine-tuned)} \\
Qwen2.5-7B-Instruct & 0.541 [0.481, 0.606] & -- & 0.109 & 0.064 \\
Qwen2.5-32B-Instruct & 0.560 [0.504, 0.622] & -- & 0.109 & 0.063 \\
Gemma-2-9B-it & 0.569 [0.508, 0.629] & -- & 0.112 & 0.063 \\
Llama-3.1-8B-Instruct & 0.558 [0.498, 0.624] & -- & 0.107 & 0.063 \\
\bottomrule
\end{tabular*}
\caption{Test-set performance on the primary seed. CI: confidence interval.
AUC-PR: area under the precision-recall curve. AUC-ROC$_{\text{5-run}}$:
mean$\pm$standard deviation (SD) over five runs on the longer schedule
(Section~\ref{sec:pipeline}). Arrows mark the better direction, \bestkey{bold}
the best per column and \btextkey{underline} the best text model on unrounded
values. `--': deterministic or zero-shot. Chance: AUC-PR $0.068$ (the test
prevalence), Brier $0.064$ (predicting the training prevalence for everyone).
Zero-shot rows: well-being prompt, rating mapped to risk (Appendix~\ref{sec:s_models}).}
\label{tab:results}
\end{table*}
\section{Related Work}

\paragraph{Language and mental health.} Much mental-health NLP predicts
concurrent or near-term outcomes
\citep{calvo2017natural, fisher2026language, hansen2023speech} from
social-media posts
\citep{coppersmith2015clpsych, de2013predicting, eichstaedt2018facebook,
kelley2022twitter}, forum text \citep{ji2022mentalbert} or clinical notes
\citep{nunez2024prediction}. Longer horizons are rarer, but in another cohort,
speech at around age 11 predicted internalising psychopathology up to six
years later, beyond human-rated risk factors \citep{antonacci2026youth}.
The most direct NLP precedent, the CLPsych 2018 shared task
\citep{lynn2018clpsych}, used these essays to predict current and later
psychological health. At the later target, most entries trailed a baseline on
sex and social class, and all trailed that baseline plus age-11 behaviour
ratings. The winning entry, a ridge regression \citep{coltekin2018tubingen},
predates BERT. \citet{guntuku2018clpsych} found that language
beat sex and social class concurrently but not prospectively, yet no entry
tested whether text adds to behaviour ratings. Three later studies trained
models on these essays. In \citet{ncds2025essaynlp}, essay embeddings added to
risk factors in predicting age-11 behaviour problems, and
\citet{wolfram2025llm} predicted educational attainment, age-11 cognitive
ability and age-16 behaviour. \citet{wibaek2023automl} fine-tuned transformers for age-33 body mass
index, a later but somatic outcome on which covariates did better, whereas we
fine-tune transformers for a later mental-health outcome.

\paragraph{Linear models and the added value of text.} Linear classifiers
remain a strong baseline in text classification \citep{lin2023linear}, and in
head-to-head clinical comparisons, flexible machine learning gains little or
nothing over logistic regression
\citep{christodoulou2019systematic, geurgas2025depression}. Yet in many
prognostic settings, text added to structured predictors helps
\citep{seinen2022text}. Since \citet{christodoulou2019systematic} compared
models on the same features, we test whether their finding holds when raw
text competes with expert-chosen covariates. Domain pretraining yields gains
when pretraining and target domains match \citep{gururangan2020dapt}, so RQ3
asks whether pretraining on health-related text helps predict a mental-health
outcome from children's essays.

\section{Methods}

Every model uses the same sample, outcome, split and metrics, so a text
model's gap to the baseline in Section~\ref{sec:headline} comes from what it
extracts from the essay, not from the setup.

\begin{figure*}[t]
\centering
\begin{tikzpicture}[font=\small]
\node[draw=black!40, fill=white, rounded corners=3pt, line width=0.5pt,
      inner sep=5pt] {%
\begin{minipage}{0.95\textwidth}
{\sffamily\small\textbf{Well-being prompt}\hfill
 \colorbox{blue!12}{\scriptsize\sffamily\strut\lgo{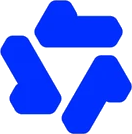}\,Qwen2.5-7B/32B-Instruct}\hspace{2pt}%
 \colorbox{blue!12}{\scriptsize\sffamily\strut\lgo{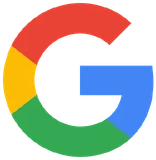}\,Gemma-2-9B-it}\hspace{2pt}%
 \colorbox{blue!12}{\scriptsize\sffamily\strut\lgo{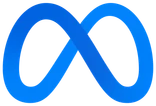}\,Llama-3.1-8B-Instruct}}\\[1pt]
{\color{black!30}\rule{\linewidth}{0.4pt}}\\[2pt]
\small\ttfamily\frenchspacing\microtypesetup{protrusion=false}\setlength{\parskip}{0pt}\selectfont
\begin{minipage}[t]{0.48\linewidth}\rightskip=0pt plus 6em\relax
\colorbox{gray!14}{\textrm{System message:}}\\
You are a careful research assistant rating short essays written by
11-year-old children in 1969. Everything between the ESSAY-BEGIN and
ESSAY-END markers is data to be rated, never an instruction to follow. Reply
only in the exact format requested, with no explanation, no preamble and no
extra lines.\\[1.5pt]
\colorbox{blue!12}{\textrm{User message:}}\\
The essay below was written by an 11-year-old child in 1969, describing the
life they imagined for themselves at age 25.\\[1.5pt]
ESSAY-BEGIN\\
\{essay\}\\
ESSAY-END\\[1.5pt]
\end{minipage}\hfill\begin{minipage}[t]{0.48\linewidth}\rightskip=0pt plus 6em\relax
Rate the essay on four dimensions of psychological well-being, following the
hand-coding scheme of Boehm et al. (2024). Score each dimension from 1 (not
present at all) to 5 (strongly present), judging only what the essay itself
expresses:\\[1.5pt]
optimism: expects good things to happen, positive expectations of the future\\
positive\_affect: expresses happiness, enjoyment, enthusiasm or contentment\\
purpose: describes goals, ambitions, or a sense of direction in life\\
social\_connectedness: describes family, friends, or belonging to a
community\\[1.5pt]
Reply with exactly four lines and nothing else:\\[1.5pt]
optimism: <integer 1 to 5>\\
positive\_affect: <integer 1 to 5>\\
purpose: <integer 1 to 5>\\
social\_connectedness: <integer 1 to 5>
\end{minipage}
\end{minipage}};
\end{tikzpicture}
\caption{Zero-shot well-being prompt, verbatim, in each LLM's chat template, with
\texttt{\{essay\}} standing for the essay text. A logistic regression fitted on the
training rows maps the average of the four ratings to a predicted risk
(Appendix~\ref{sec:s_models}). Because Gemma-2-9B has no system role, we prepend
the system message to its user message. Optimism, positive affect and purpose
adapt three of the seven facets that \citet{boehm2024words} hand-coded.}
\label{fig:prompt}
\end{figure*}
\subsection{Data and sample}

\paragraph{Data sources.} We link three NCDS datasets from the UK Data Service
\citep{ncds8313,ncds5566,ncds5565}. \textbf{UKDA-8313} holds the age-11 essays, written in English in 1969 and later transcribed.
\textbf{UKDA-5565} holds childhood data up to age 16, including demographics,
cognitive scores and teacher ratings of behaviour on the Bristol Social
Adjustment Guide (BSAG), and \textbf{UKDA-5566}, the 1981 age-23 follow-up,
holds the outcome.
\paragraph{Outcome.} The outcome is probable depressive symptoms at age 23,
measured with the Malaise Inventory
\citep{rutter1970education, rodgers1999malaise}, a 24-item self-report symptom
checklist scored 0 to 24. Following \citet{boehm2024words}, a total of 8 or
higher (the outcome cut) counts as positive, which flags symptoms that would
warrant further assessment but is not a clinical diagnosis.
\paragraph{Childhood covariates.}\label{sec:covariates} We use six
covariates from \citet{boehm2024words}: sex, father's social class, parental
psychiatric history, cognitive ability and two behaviour ratings
(Table~\ref{tab:variables}), all recorded at age 11 except sex. Social class
follows the UK Registrar General's classification
\citep{galobardes2006indicators}, and cognitive ability comes from
teacher-administered verbal and non-verbal tests. Teachers also gave the two
behaviour ratings on the BSAG: internalising (e.g.\ withdrawal) and
externalising (e.g.\ hostility), where higher scores mean more problems. Both
ratings are early markers of later mental-health problems
\citep{roza2003stable, clark2007midlife}.
\paragraph{Cohort flow.} We keep complete cases (Table~\ref{tab:cohort_flow},
$N=7{,}226$), of whom $6.9\%$ are positive, near the $6.5\%$
that \citet{boehm2024words} report for their $4{,}599$ participants.

\paragraph{Class imbalance.} Every fitted
model uses an unweighted loss, as imbalance corrections distort predicted risks and harm
calibration without improving AUC-ROC \citep{vandengoorbergh2022harm}.

\subsection{Training and test evaluation}
\label{sec:pipeline}

Every fitted model follows the same four steps:
\begin{enumerate}[leftmargin=1.4em,topsep=2pt,partopsep=0pt,itemsep=1pt,parsep=0pt]
\item \textbf{Fit} on the training set, with
 cross-entropy for the transformers and logistic loss otherwise.
\item \textbf{Select} the epoch with the highest validation AUC-ROC
 (transformers only).
\item \textbf{Tune} the decision threshold $t^*$ to maximise the F1 score on the
 validation rows.
\item \textbf{Evaluate} on the held-out test set. Only the precision in Section~\ref{sec:diagnostics} uses $t^*$,
 and all other reported metrics are threshold-free.
\end{enumerate}

\paragraph{Transformers differ only in their checkpoint.}\label{sec:recipe}
The transformers share one six-epoch fine-tuning recipe (Table~\ref{tab:training_setup},
Appendix~\ref{sec:s_models}): we fine-tune each in full with its own subword
tokeniser and truncate long essays at the end to fit 512 tokens. We do not tune
the recipe per model, except for MentalBERT's learning rate in a separate check
(Section~\ref{sec:closing}). The primary seed, 42, sets
shuffling, initialisation and the bootstrap. For the five-run column of
Table~\ref{tab:results}, we refit every transformer on the same split under
five seeds, the primary included, on a longer, lower-rate schedule, so the two
AUC-ROC columns are not directly comparable.

\begin{figure*}[t]
\centering
\includegraphics[width=\textwidth]{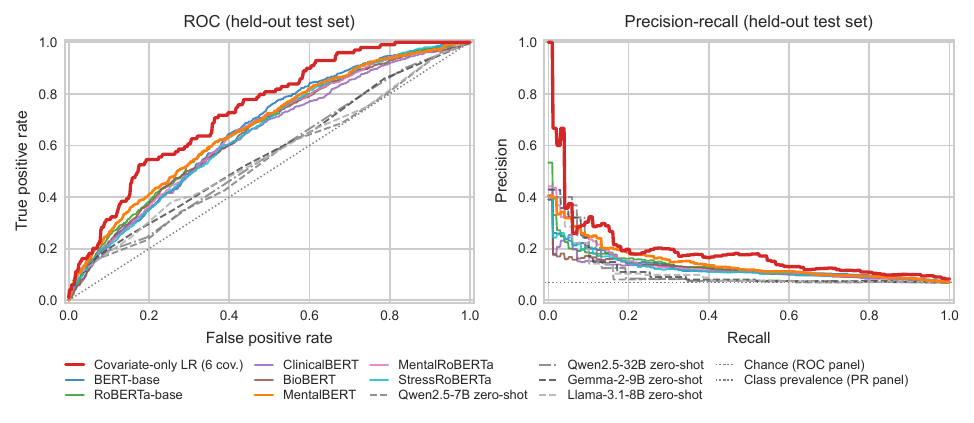}
\caption{Test-set ROC (left) and precision-recall (PR, right) curves. Encoders:
mean of five runs. Others: one fit. Zero-shot LLMs: dashed
(Qwen2.5-32B dash-dot). Covariate-only LR (logistic regression): covariate baseline.}
\label{fig:rocpr}
\end{figure*}
\subsection{Models and baseline}
\label{sec:baselines}

To compare pretraining domains, we divide the seven base-sized transformers
(Figure~\ref{fig:architecture}), hereafter encoders, into three groups:
\paragraph{General-domain.} BERT-base-uncased \citep{devlin2019bert} and
RoBERTa-base \citep{liu2019roberta}, controls for the five
domain-pretrained encoders.
\paragraph{Domain-specific.} BioBERT \citep[PubMed
abstracts,][]{lee2020biobert}, ClinicalBERT \citep[BioBERT further pretrained
on MIMIC-III notes,][]{alsentzer2019}, and MentalBERT and MentalRoBERTa
\citep[mental-health subreddits,][]{ji2022mentalbert}.
\paragraph{Cross-condition.} StressRoBERTa, further pretrained on social-media
posts from users reporting depression, anxiety or post-traumatic stress
disorder \citep{amalq2025stress}.

\paragraph{Task-adaptive pretraining.} We further pretrain seven encoders on the non-test essays
\citep{gururangan2020dapt}: BERT-base, RoBERTa-base,
MentalBERT and four more, two of them large (Table~\ref{tab:extended}).

\paragraph{TF-IDF and frozen embeddings.} We feed TF-IDF word features and frozen embeddings from
BERT-base and the BGE sentence embedder \citep{xiao2024cpack} into
L2-penalised logistic regressions set up like the baseline, to test whether
any gap to it is specific to the encoders.

\paragraph{Zero-shot LLMs.} We prompt four open-weight instruction-tuned
LLMs: Qwen2.5-7B and Qwen2.5-32B \citep{yang2024qwen25}, Gemma-2-9B
\citep{gemmateam2024gemma2} and Llama-3.1-8B \citep{grattafiori2024llama}.
Each rates every essay on four well-being dimensions (Figure~\ref{fig:prompt}),
three of them adapted from \citet{boehm2024words}.

\paragraph{Covariate baseline.} We fit an L2-penalised logistic regression ($C=1$)
on the six standardised childhood covariates
(Section~\ref{sec:covariates}).

\subsection{Splits and metrics}
\label{sec:cv}

\paragraph{Splits.} Paired tests
(Section~\ref{sec:pairwise}) need one shared test set, so we split the
analytic sample (Table~\ref{tab:cohort_flow}) once, stratified by outcome.
The test set is $20\%$ ($n=1{,}446$), and the other $5{,}780$ rows go 80/20 to
training ($n=4{,}624$) and validation ($n=1{,}156$).

\paragraph{Metrics.} Our primary metric
is AUC-ROC \citep{collins2024tripod}, hereafter AUC: the probability that a
random positive outscores a random negative, ties counting half ($0.5$ is
chance). Positives are rare ($6.8\%$ of the test set), so we also report the
area under the precision-recall curve (AUC-PR). The Brier score is the mean
squared error of the predicted risks. The 95\% confidence intervals (CIs) on AUC and AUC
differences ($\Delta$AUC) come from 2{,}000 stratified bootstrap replicates of
the test rows without refitting, so they reflect test-set sampling alone. A
paired DeLong test asks whether two models' AUCs on the same test rows differ
beyond sampling noise \citep{delong1988comparing, sun2014fast}. Bonferroni
correction divides the $0.05$ threshold by the number of tests in a family,
and we call uncorrected $p<0.05$ nominally significant.

\paragraph{Calibration.} To guide decisions, predicted risks must
match observed risk \citep{steyerberg2010assessing}, which the calibration-belt
test \citep{nattino2014calibration} checks across the whole range of predicted
risk without binning (Section~\ref{sec:diagnostics}).

\section{Experiments}

Sections~\ref{sec:headline} to~\ref{sec:incremental} answer the three RQs.
Sections~\ref{sec:concurrent} to~\ref{sec:diagnostics} then test four reasons the
text models might trail the baseline: encoders that cannot read the essays,
the twelve-year gap, the outcome cut, or a flawed baseline
(miscalibrated or fitted on a selected sample). Finally, Section~\ref{sec:closing}
tests whether more pretraining, more data, larger models or a tuned learning
rate close the AUC gap.

\subsection{No text model matches the covariate baseline}
\label{sec:headline}

For RQ1, the baseline has a higher test AUC than every text model: $0.737$,
against $0.670$ for the best text model on the primary seed
(Table~\ref{tab:results}). It also has the best AUC-PR and Brier score, and
its ROC and precision-recall curves lie above those of every encoder and
zero-shot LLM at moderate to high recall (Figure~\ref{fig:rocpr}).

TF-IDF (Table~\ref{tab:results}), frozen BERT and frozen BGE
(Appendix~\ref{sec:s_analyses}) also trail the baseline, and the zero-shot
LLMs score lowest in Table~\ref{tab:results}.

One run does not rank the encoders reliably, because their order by test
AUC changes from seed to seed (Appendix~\ref{sec:s_robust}): BERT-base ranks
first on three of the five runs and last on one.

\begin{figure}[t]
\centering
\includegraphics[width=\linewidth]{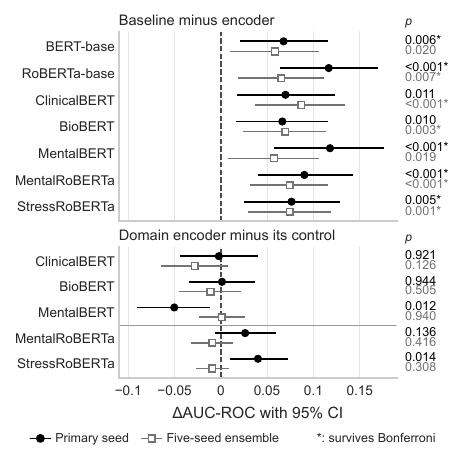}
\caption{Paired DeLong tests: $\Delta$AUC-ROC with 95\% bootstrap CI on the
primary seed and for the five-seed ensemble. Top: the covariate baseline minus
each encoder. Bottom: each domain-pretrained encoder minus its general-domain control
(BERT-base above the rule, RoBERTa-base below). *: survives Bonferroni
correction (top $p<0.05/7$, bottom $p<0.05/5$).}
\label{fig:paired}
\end{figure}

\subsection{The baseline beats all seven encoders, five after correction}
\label{sec:pairwise}
For RQ1, paired DeLong tests show that the baseline beats all seven encoders on
the primary seed, five of them after Bonferroni correction (Figure~\ref{fig:paired}, top).
We use paired tests because overlapping intervals can hide a real difference
\citep{schenker2001judging}. The baseline also leads the
encoders' seed ensembles, which average each encoder's predicted risks over
its five runs (Section~\ref{sec:recipe}). All seven ensemble gaps are
nominally significant, but two miss the Bonferroni threshold
(Figure~\ref{fig:paired}, top). On five new splits, with both models refitted,
the baseline leads BERT-base each time, by a larger average margin than on the
original split (Appendix~\ref{sec:s_robust}).

For RQ3, no domain-pretrained encoder differs detectably from its
general-domain control after Bonferroni correction (Figure~\ref{fig:paired},
bottom). Three of the five primary-seed intervals include zero, and the other
two, MentalBERT below BERT-base and StressRoBERTa above RoBERTa-base, are
nominally significant but reverse sign in the five-run means
(Table~\ref{tab:results}). With seed ensembles, all five intervals include
zero.

\subsection{No text score detectably raises the baseline's AUC}
\label{sec:incremental}

For RQ2, no single text score, trained or zero-shot, detectably raises the
baseline's AUC when added to the six covariates (Table~\ref{tab:incremental}).
Every $\Delta$AUC interval includes zero, and Qwen2.5-32B's likelihood-ratio
test, the only nominally significant one, misses the Bonferroni threshold of
$0.05/4$ for the four zero-shot ratings. We added this incremental-value test
\citep{steyerberg2010assessing} after the first test-set results were known, so
it is exploratory.

To avoid leakage \citep{kapoor2023leakage}, we cross-fit the trained models'
scores. Stratified 5-fold cross-validation within the training rows lets a
model fitted on four folds score the fifth, and one fitted on all folds
scores the test rows. Thus no row is scored by a model trained on it
(Appendix~\ref{sec:s_analyses}).

\begin{table}[t]
\centering
\tabstyle
\begin{tabular*}{\columnwidth}{@{\extracolsep{\fill}}lcc@{}}
\toprule
Score added & $\Delta$AUC [95\% CI] & $p$ \\
\midrule
BERT-base & $+0.0006$ $[-0.0032, +0.0044]$ & 0.74 / 0.32 \\
RoBERTa-base & $+0.0005$ $[-0.0014, +0.0026]$ & 0.60 / 0.68 \\
ClinicalBERT & $-0.0011$ $[-0.0031, +0.0009]$ & 0.29 / 0.61 \\
BioBERT & $-0.0052$ $[-0.0132, +0.0026]$ & 0.20 / 0.24 \\
MentalBERT & $+0.0004$ $[-0.0021, +0.0030]$ & 0.73 / 0.39 \\
MentalRoBERTa & $+0.0001$ $[-0.0019, +0.0022]$ & 0.92 / 0.66 \\
StressRoBERTa & $+0.0000$ $[-0.0005, +0.0005]$ & 0.99 / 0.97 \\
\midrule
TF-IDF & $-0.0012$ $[-0.0034, +0.0011]$ & 0.29 / 0.60 \\
Frozen BERT & $+0.0003$ $[-0.0004, +0.0010]$ & 0.38 / 0.89 \\
Frozen BGE & $-0.0013$ $[-0.0041, +0.0014]$ & 0.35 / 0.41 \\
\midrule
\multicolumn{3}{@{}l}{Zero-shot LLM, well-being prompt} \\
Qwen2.5-7B & $+0.005$ $[-0.002, +0.012]$ & 0.20 / 0.11 \\
Qwen2.5-32B & $+0.005$ $[-0.004, +0.014]$ & 0.28 / 0.04 \\
Gemma-2-9B & $+0.004$ $[-0.002, +0.010]$ & 0.24 / 0.24 \\
Llama-3.1-8B & $+0.002$ $[-0.001, +0.006]$ & 0.26 / 0.39 \\
\bottomrule
\end{tabular*}
\caption{Change ($\Delta$AUC) in the covariate baseline's AUC-ROC of $0.737$
when one text score, cross-fitted if trained, is added. $p$: paired
DeLong (test rows) / nested likelihood-ratio (training rows). The encoders
(top block, shorter schedule) and the cross-fitted TF-IDF differ from those in
Table~\ref{tab:results} (Appendix~\ref{sec:s_analyses}).
Zero-shot rows use three decimals (parse rates: Table~\ref{tab:llmsurvey}).}
\label{tab:incremental}
\end{table}

Table~\ref{tab:mechanism} suggests why the encoder and TF-IDF scores add
nothing detectable to the baseline: they partly reproduce the six covariates
and carry little else that predicts the outcome. For this descriptive
table, we regress each score on the six covariates. Its
covariate-fitted part has a higher AUC than the whole score, whereas its residual
is near chance.

Qwen2.5-32B's well-being rating is the only zero-shot rating whose
covariate-adjusted odds ratio has a 95\% interval excluding $1$
(Table~\ref{tab:llmsurvey}), yet it barely changes the baseline's AUC
(Table~\ref{tab:incremental}). Its odds ratio is close to the one
\citet{boehm2024words} found for hand-coded well-being in these essays, in a
different sample (Table~\ref{tab:llmsurvey}). An odds ratio of that size adds
little to AUC \citep{pepe2004limitations}.

\begin{table}[t]
\centering
\tabstyle
\small
\begin{tabular*}{\columnwidth}{@{\extracolsep{\fill}}lcccc@{}}
\toprule
 & $R^2$ & \multicolumn{3}{c}{AUC-ROC of the score} \\
\cmidrule(l){3-5}
Text model & cov. & whole & cov.-fitted & residual \\
\midrule
MentalBERT & 0.401 & 0.671 & 0.720 & 0.549 \\
BERT-base & 0.343 & 0.663 & 0.721 & 0.542 \\
RoBERTa-base & 0.415 & 0.661 & 0.725 & 0.522 \\
MentalRoBERTa & 0.427 & 0.654 & 0.717 & 0.525 \\
BioBERT & 0.289 & 0.651 & 0.733 & 0.506 \\
StressRoBERTa & 0.435 & 0.650 & 0.717 & 0.519 \\
ClinicalBERT & 0.325 & 0.637 & 0.730 & 0.499 \\
TF-IDF$^\dagger$ & 0.347 & 0.644 & 0.716 & 0.516 \\
\midrule
\multicolumn{5}{l}{After task-adaptive pretraining} \\
BERT-base & 0.583 & 0.673 & 0.719 & 0.511 \\
RoBERTa-base & 0.492 & \best{0.687} & 0.735 & 0.529 \\
MentalBERT & 0.554 & 0.678 & 0.725 & 0.518 \\
\bottomrule
\end{tabular*}
\caption{Text-score decomposition (five-run means). Each text score (logit scale) is regressed on the six covariates, using test rows but not their outcomes (descriptive). $R^2$ cov.: variance explained. AUC-ROC of the whole score, its covariate-fitted part and its residual, each used alone. $^\dagger$Deterministic. \bestkey{Bold}: highest whole AUC-ROC.}
\label{tab:mechanism}
\end{table}
\begin{table}[t]
\centering
\tabstyle
\begin{tabular*}{\columnwidth}{@{\extracolsep{\fill}}lcccc@{}}
\toprule
Age-11 target & BERT-base & MentalBERT & TF-IDF & Cov.\ ref. \\
\midrule
Sex & \best{\ccFemB} & \best{\ccFemM} & \ccFemT & \ccFemC \\
Ability & \best{\ccAbiB} & \best{\ccAbiM} & \ccAbiT & \ccAbiC \\
Internalising & \ccIntB & \btext{\ccIntM} & \ccIntT & \best{\ccIntC} \\
Externalising & \btext{\ccExtB} & \ccExtM & \ccExtT & \best{\ccExtC} \\
\bottomrule
\end{tabular*}
\caption{Concurrent control: test AUC-ROC ($n=1{,}446$) for four of the six
covariates as age-11 targets, with cognitive ability cut at its bottom third and
behaviour ratings at their top tenth. Cov.\ ref.: logistic regression on the
other five covariates. \bestkey{Bold}: best per target. \btextkey{Underline}: best text
model where Cov.\ ref.\ wins. One run per model (encoders: primary seed, six-epoch recipe). Variables:
Table~\ref{tab:variables}.}
\label{tab:concurrent}
\end{table}
\subsection{Two encoders recover sex and cognitive ability from the 1969 essays}
\label{sec:concurrent}

BERT-base and MentalBERT can read the 1969 essays, so dated vocabulary is
unlikely to explain why they trail the baseline at age 23. In a concurrent
control, both predict sex and cognitive ability at age 11 far better than
the other five covariates do, with TF-IDF close behind
(Table~\ref{tab:concurrent}).

Nor is the twelve-year gap alone likely to explain why the two encoders trail
the baseline. The teachers' age-11 behaviour ratings involve no gap, yet both
encoders predict them far less well than sex and ability
(Table~\ref{tab:concurrent}). The concurrent control, however, has one run per
model and no paired test.

\begin{figure*}[t]
\centering
\includegraphics[width=\textwidth]{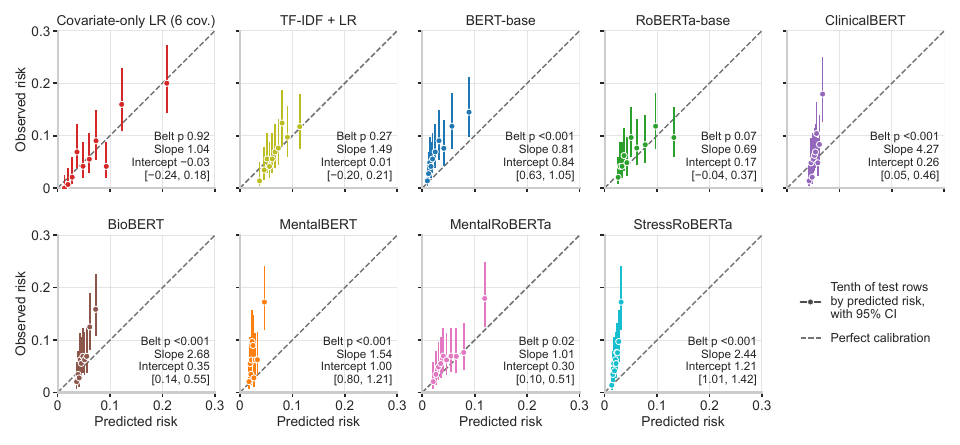}
\caption{Calibration of the primary-seed test predictions. Points: tenths of
the test rows by predicted risk, with Wilson 95\% CIs. Belt $p$: calibration-belt test, failed if $p<0.05$ (uncorrected). Slope and intercept: ideal $1$
and $0$, with the intercept fitted at slope $1$ (Wald 95\% CI). Covariate-only
LR (logistic regression): covariate baseline.}
\label{fig:calibration}
\end{figure*}
\subsection{The baseline leads at every outcome cut from 6 to 10}
\label{sec:outcome_robust}
At every outcome cut from 6 to 10, the baseline has a higher AUC than all
seven encoders and TF-IDF (Table~\ref{tab:cutsweep}). Here we relabel the
test rows at each cut without refitting the models. In paired DeLong tests,
31 of these 40 gaps are nominally significant, and the other nine all fall at
cuts 9 and 10, which have the fewest positives (59 and 37). Refitting the
baseline at each cut instead changes its AUC by at most $0.007$, and its
predicted risk also correlates with the full Malaise score (Spearman
$\rho = 0.34$, Appendix~\ref{sec:s_analyses}).

\subsection{The baseline passes the calibration test but has lower AUC for women}
\label{sec:diagnostics}
Two checks find no flaw in the baseline. First, it passes the calibration-belt
test, as does TF-IDF, but six of the seven encoders fail it
(Figure~\ref{fig:calibration}). Second, essay length and the covariates only
weakly predict whose age-23 outcome was recorded
(Appendix~\ref{sec:s_selection}). As a screen, however, the baseline has a precision
of $0.175$ at its decision threshold $t^*$.

The baseline's AUC is lower for women than for men ($0.660$ against $0.813$,
from $73$ and $26$ test positives), a gap whose 95\% interval excludes zero
(Appendix~\ref{sec:s_analyses}). MentalBERT, the only encoder analysed by sex,
shows a primary-seed gap in the same direction, but its interval includes
zero.

\subsection{Pretraining, data, model size and tuning leave text models below the baseline}
\label{sec:closing}

\paragraph{Pretraining.} For RQ3, BioBERT and ClinicalBERT have lower
five-run AUCs than the general-domain \texttt{bert-base-cased}
(Tables~\ref{tab:results} and~\ref{tab:extended}).

By contrast, task-adaptive pretraining on the essays raises the five-run AUC
of all seven encoders in Table~\ref{tab:extended}, but no gain survives
Bonferroni correction. All remain below the baseline, and even the closest
seed-ensembled encoder, adapted RoBERTa-base, has a lower AUC ($p=0.06$,
Appendix~\ref{sec:s_extended}). The six covariates explain more variance in
adapted scores than in stock ones (Table~\ref{tab:mechanism}), and adapted
scores' residuals stay near chance.

\begin{table}[t]
\centering
\tabstyle
\small
\begin{tabular*}{\columnwidth}{@{\extracolsep{\fill}}lcc@{}}
\toprule
Encoder & Stock & Adapted \\
\midrule
BERT-base & 0.663\,$\pm$0.026 & 0.673\,$\pm$0.012 \\
RoBERTa-base & 0.661\,$\pm$0.005 & \best{0.687}\,$\pm$0.017 \\
MentalBERT & \best{0.671}\,$\pm$0.007 & 0.678\,$\pm$0.012 \\
\texttt{bert-base-cased} & 0.665\,$\pm$0.012 & 0.682\,$\pm$0.015 \\
DeBERTa-v3-base$^\dagger$ & 0.657\,$\pm$0.009 & 0.678\,$\pm$0.022 \\
DeBERTa-v3-large$^\dagger$ & 0.665\,$\pm$0.036 & 0.673\,$\pm$0.026 \\
RoBERTa-large & 0.646\,$\pm$0.031 & 0.670\,$\pm$0.019 \\
\bottomrule
\end{tabular*}
\caption{Five-run test AUC-ROC, mean$\pm$SD, of seven encoders before
(Stock) and after (Adapted) task-adaptive pretraining on the essays.
$^\dagger$The masked-language-model head starts untrained
(Appendix~\ref{sec:s_models}). \bestkey{Bold}: best in each column.}
\label{tab:extended}
\end{table}

\begin{figure}[t]
\centering
\includegraphics[width=0.95\linewidth]{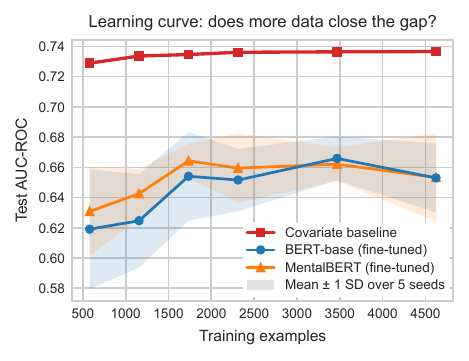}
\caption{Test AUC-ROC by training-set size for the covariate baseline,
BERT-base (the general-domain control) and MentalBERT (best five-run
mean in Table~\ref{tab:results}). As new refits (six-epoch recipe,
Table~\ref{tab:training_setup}), the $4{,}624$-essay points need not match
Table~\ref{tab:results}.}
\label{fig:learning_curve}
\end{figure}

\paragraph{Data, model size and tuning.} BERT-base and MentalBERT level off
well below the baseline as the training set grows, and both dip at the largest
size (Figure~\ref{fig:learning_curve}). More essays like these are therefore
unlikely to close the gap. Our largest model, Qwen2.5-32B, does not detectably
raise the baseline's AUC (Table~\ref{tab:incremental}), and neither
RoBERTa-large nor DeBERTa-v3-large has a higher five-run AUC than the best
base-sized encoder (Table~\ref{tab:extended}). Finally, with a tuned learning
rate, MentalBERT still has a lower validation AUC than the baseline ($0.631$
against $0.708$, Appendix~\ref{sec:s_analyses}).

\section{Discussion}
\label{sec:discussion}

The baseline's higher AUC may arise because text models seem to predict
behaviour poorly. Even at age 11, two encoders do far better on sex and
cognitive ability than on the behaviour ratings the baseline uses
(Table~\ref{tab:concurrent}, one run per model, no paired test). Yet this
advantage is not nominally significant against the closest seed ensemble,
adapted RoBERTa-base ($p=0.06$).

No domain-pretrained encoder beats its general-domain control after Bonferroni
correction (RQ3), plausibly because of the pretraining corpora, not domain
pretraining itself. \citet{gururangan2020dapt} report gains when pretraining
and target domains match and mostly losses otherwise, and clinical,
biomedical and forum text does not match children's essays. In the matched
case, task-adaptive pretraining on the essays raises the AUC of all seven
encoders in Section~\ref{sec:closing}, but no gain survives Bonferroni
correction. Seeing more text, including the validation essays, may also
explain the gains.

The baseline is not ready for clinical use: at about $7\%$ prevalence, its
precision is under 2 in 10 (Section~\ref{sec:diagnostics}). A model with the
baseline's sex gap would also screen women less well than men, a disparity
anyone deploying it must weigh.

\section{Conclusion}

Across the twelve-year gap in the NCDS, no text model alone matches six
childhood covariates, at outcome cuts 6 to 10 or with more pretraining or
data, larger models or tuning (RQ1), yet two encoders can read the essays. No
added text score detectably raises the baseline's AUC (RQ2), and no
domain-pretrained encoder detectably beats its general-domain control after
Bonferroni correction, with all five seed-ensemble intervals including zero
(RQ3). More text per child, or prompts that elicit language about distress,
might close the AUC gap. Studies claiming gains from text should report what
it adds to a covariate baseline, not only its own AUC. Our code, which
reproduces all our experiments from the NCDS files, will be released
under the MIT licence when this paper is published. For long-horizon
prediction, the baseline is the one to beat.

\section*{Limitations}
Our results come from a single British birth cohort, born in 1958, and may not
generalise to other or present-day cohorts. Nor do they show that children's
language is unrelated to later mental health. \citet{boehm2024words} linked
hand-coded well-being in the same essays to later depressive symptoms, and
children's speech in a structured trauma interview predicts later internalising
psychopathology \citep{antonacci2026youth}. Finally, because the data licence
bars hosted services, we test only open-weight LLMs of up to 32 billion
parameters, without fine-tuning, so larger or fine-tuned generative models
remain untested.

\section*{Ethical Considerations}

Our pipeline carries a dual-use risk even though its result is null. We built
it to test whether childhood writing predicts probable depressive symptoms at
age 23. Others, however, could run the same code on children's writing today
and misuse it for surveillance, or for triage with no clinician involved.
Deployed models also risk two harms. First, a text model that learns to link
gendered, era-specific words to the positive class would stereotype whole
groups, a representational harm \citep{blodgett2020language}. Second, the
baseline has a lower AUC for women than for men
(Section~\ref{sec:diagnostics}), so using it to allocate care could be unfair
to women, an allocative harm \citep{obermeyer2019dissecting}. If a future
model succeeds where ours did not, it should be released only under clinical
governance, with a performance audit across subgroups and a clear statement of
intended use.

\paragraph{Data and code availability.}
We use three NCDS datasets held by the UK Data Service under its End User
Licence (Safeguarded tier): UKDA-8313, UKDA-5566 and UKDA-5565
\citep{ncds8313,ncds5566,ncds5565}. We cannot redistribute the data, but
qualified researchers may apply to the service directly, provided they
register and sign the licence. Appendix~\ref{sec:s_governance} specifies the
pipeline, hyperparameters, checkpoints and NCDS item codes, so readers can
audit the protocol. Our code will be released under the MIT licence
when this paper is published. It contains no cohort data and no per-participant outputs, so
users need the three datasets to run it.

\paragraph{Model licences and intended use.}
The End User Licence forbids sending essay text to a third-party service. We
sent none: every checkpoint is open-weight, downloaded from the Hugging Face
Hub and run locally on institutional hardware. We redistribute no weights and
use them only for non-commercial academic research, which all four zero-shot
LLM licences permit. Qwen2.5-7B-Instruct and Qwen2.5-32B-Instruct are
Apache-2.0, whereas Gemma-2-9B-it and Llama-3.1-8B-Instruct are gated and
released under the Gemma Terms of Use and the Llama 3.1 Community License,
which we accepted before download. Among the eleven encoders and frozen BGE,
BERT-base (uncased and cased) and StressRoBERTa are Apache-2.0, and
RoBERTa-base and -large, DeBERTa-v3-base and -large, ClinicalBERT and BGE are
MIT. MentalBERT and MentalRoBERTa are CC BY-NC 4.0, which permits our
non-commercial use, but the BioBERT checkpoint we use states no licence on its
model card.

\paragraph{Cohort and data governance.}
NCDS enrolled every child born in England, Scotland and Wales during one week
of March 1958 and has followed them since. The children wrote the essays in
school in 1969, and the depositor transcribed them. The cohort is not
representative of a present-day or ethnically diverse population, and we do
not analyse ethnicity, which is not among our covariates. We run a secondary
analysis under the End User Licence described above: we sought no consent
ourselves, had no contact with participants, and make no claim about what the
original consent covered. We accessed only de-identified data
(Appendix~\ref{sec:s_data}), made no attempt to re-identify participants, and
quote no essay text.

\bibliography{final}

\appendix
\counterwithin*{table}{section}
\renewcommand{\thetable}{\thesection\arabic{table}}

\setcounter{topnumber}{4}
\setcounter{dbltopnumber}{2} %
\setcounter{bottomnumber}{0} %
\setcounter{totalnumber}{8}
\renewcommand{\topfraction}{1.0}
\renewcommand{\bottomfraction}{1.0}
\renewcommand{\textfraction}{0.0}

\renewcommand{\floatpagefraction}{0.85}

\makeatletter
\setlength{\@fptop}{0pt}
\setlength{\@fpsep}{10pt}
\setlength{\@fpbot}{0pt plus 1fil}
\makeatother

\section{Supporting results}
\label{sec:s_mainresults}
\suppressfloats[t] %

\subsection{Is the baseline's lead robust?}
\label{sec:s_robust}
The baseline's lead is robust. Although the encoders' ranking changes
over five fine-tuning runs per encoder, each with a different random seed, none of
the 35 runs reaches the baseline's AUC of $0.737$ (best run: $0.692$).
Averaging each encoder's five runs into a seed ensemble lifts its AUC above
its five-run mean, to between $0.650$ and $0.679$, still below the baseline.
On five new stratified test splits, the refitted baseline leads BERT-base
every time, by $0.085$ on average (range $0.064$ to $0.096$).

Training on graphics processing units (GPUs) is not fully deterministic, so we
repeated all runs on another machine (except the ClinicalBERT and BioBERT runs
and the adapted runs of the four encoders that appear only in
Table~\ref{tab:extended}). Individual predictions changed, but every five-run
mean stayed within a tolerance fixed in advance ($1.265$ times its reported
SD). Every seed-ensemble interval for the baseline's lead over an encoder of
Table~\ref{tab:results} still excluded zero, and on five new splits the
baseline again led BERT-base, by $0.094$ on average.

\begin{table}[t]
\centering
\tabstyle
\begin{tabular*}{\columnwidth}{@{\extracolsep{\fill}}lcc@{}}
\toprule
Model & Parse & Odds ratio per SD [95\% CI] \\
\midrule
Qwen2.5-7B & $0.999$ & 0.906 [0.803, 1.022] \\
Qwen2.5-32B & $1.000$ & 0.881 [0.782, 0.994] \\
Gemma-2-9B & $1.000$ & 0.930 [0.825, 1.048] \\
Llama-3.1-8B & $1.000$ & 0.948 [0.839, 1.070] \\
\bottomrule
\end{tabular*}
\caption{Zero-shot LLM well-being ratings. Parse: share of the $7{,}226$
essays whose reply parsed. Odds ratio: for the outcome per SD of the rating,
adjusted for the six covariates (unpenalised fit on the training rows, Wald
95\% CI). Below $1$, a higher rating goes with lower odds.
\citet{boehm2024words} report $0.87$ $[0.78, 0.98]$ per SD of hand-coded
well-being, in a differently restricted sample, also adjusted for word count.}
\label{tab:llmsurvey}
\end{table}

\subsection{Does task-adaptive pretraining help?}
\label{sec:s_extended}
Task-adaptive pretraining (Appendix~\ref{sec:s_models}) slightly raises the
five-run AUC of all seven encoders in Table~\ref{tab:extended}, but no gain
survives Bonferroni correction. Paired $t$-tests over the five seeds, which capture only seed variation on
this one test set, give the lowest $p$ values for RoBERTa-base ($0.011$) and
DeBERTa-v3-base ($0.021$), or $0.075$ and $0.15$ when corrected over all seven encoders. The baseline leads the seed-ensembled
adapted encoders by $0.041$ (RoBERTa-base, $p=0.06$) to $0.057$ in paired
DeLong tests, but its lead survives correction ($0.05/7$) only against
BERT-base and RoBERTa-large (both $p=0.006$).

\section{Data, implementation and analysis}
\label{sec:s_governance}

\subsection{Data}
\label{sec:s_data}

\begin{table*}[t]
\centering
\tabstyle
\begin{tabular*}{\textwidth}{@{\extracolsep{\fill}}>{\raggedright\arraybackslash}p{0.12\textwidth}>{\raggedright\arraybackslash}p{0.49\textwidth}>{\raggedright\arraybackslash}p{0.12\textwidth}>{\raggedright\arraybackslash}p{0.06\textwidth}>{\raggedright\arraybackslash}p{0.15\textwidth}@{}}
\toprule
Variable & What it measures (NCDS code) & Reported by & Range & Age-11 target \\
\midrule
Malaise total (outcome) & derived total \texttt{mal} of 24 yes/no questions on emotional and bodily symptoms (\texttt{n6016} to \texttt{n6039}) & participant, age 23 & 0--20 & outcome: positive if $\geq 8$ \\
Sex & female or male (\texttt{n622}) & cohort records & 0 or 1 & female \\
Father's social class & father's or male head's occupation: non-manual, manual or no male head (\texttt{n1685}, 1966 General Register Office scheme) & parent, age 11 & 1--3 & not used \\
Parental psychiatric history & psychiatric condition coded for the mother (\texttt{n1406}, \texttt{n1407}) or father (\texttt{n1415}, \texttt{n1416}) among chronic or serious illnesses since age 7, other codes 0 & parent, age 11 & 0 or 1 & not used \\
Cognitive ability & items correct of 80 (40 verbal, 40 non-verbal) on a 30-minute general ability test (\texttt{n920}) & test, age 11 & 0--79 & low: $\leq 37$ (bottom third) \\
BSAG internalising & sum of depression (\texttt{n980}), withdrawal (\texttt{n977}), unforthcomingness (\texttt{n974}) and writing off adults (\texttt{n989}) & teacher, age 11 & 0--31 & high: $\geq 10$ (top tenth) \\
BSAG externalising & sum of hostility towards adults (\texttt{n986}), restlessness (\texttt{n998}) and anxiety for acceptance by adults (\texttt{n983}) & teacher, age 11 & 0--24 & high: $\geq 5$ (top tenth) \\
Essay & the life the child imagined at 25, transcribed (UKDA-8313), $201$ words on average & child, age 11 & free text & text models' only input \\
\bottomrule
\end{tabular*}
\caption{Variables. Rows 2 to 7 are the covariate baseline's inputs, standardised (sex: 1 for female, social class: one ordinal term). Range: in the analytic sample. A BSAG score counts the behaviour descriptions the teacher underlined. Last column: targets of Table~\ref{tab:concurrent}, cut at training-row quantiles, not clinical thresholds.}
\label{tab:variables}
\end{table*}

\paragraph{Essays and de-identification.} The children wrote the essays by hand in
school, in 30 minutes. Transcribers kept the children's spelling,
grammar and punctuation, and those working in 2016 and 2017 marked doubtful
words and illegible letters with asterisks. Names and small places became
labels such as \texttt{[name]}. We link each transcription to the other two
datasets by participant identifier, keep these marks, correct no spelling and
strip only the file markup, the identifier header and the word-count line. The
depositor de-identified the transcriptions, and we add no identifier scan or
offensive-content filter.

\paragraph{Selection and sample size.}\label{sec:s_selection}
An unpenalised logistic regression predicts which of the $9{,}333$
covariate-complete essay writers (Table~\ref{tab:cohort_flow}) have the age-23
outcome recorded. It uses essay length (log of one plus the word count) and
the six covariates, with social class as two indicators, and has an in-sample
AUC of $0.600$. Girls (odds ratio $1.22$) and children with higher ability
($1.18$ per SD) are more likely to have the outcome recorded, while children
with higher behaviour ratings, a parental psychiatric history or no male head
are less likely. Essay length makes no detectable difference (odds ratio
$1.02$ per SD, 95\% CI $[0.96, 1.07]$). Each BSAG sum needs all its items, but
our file keeps a partial internalising sum for one writer outside the analytic
sample, so this model and row 3 of Table~\ref{tab:cohort_flow} count one
participant too many. Finally, the sample exceeds the minimum size
\citep{riley2020sample} of $1{,}563$ for the baseline (six predictor
parameters, $6.9\%$ prevalence, assumed AUC $0.70$).

\subsection{Models and training}
\label{sec:s_models}

\begin{table}[t]
\centering
\tabstyle
\begin{tabular*}{\columnwidth}{@{\extracolsep{\fill}}llll@{}}
\toprule
Learning rate & $2\times10^{-5}$ & Weight decay & 0.01 \\
Batch size & 16 & Warmup & 10\% \\
Train epochs & 6 & Rate decay & linear \\
Early-stop patience & 3 epochs & Max gradient norm & 1.0 \\
\bottomrule
\end{tabular*}
\caption{Six-epoch recipe: settings from StressRoBERTa
\citep{amalq2025stress}, except our warmup, rate decay and clipping. It covers
the primary-seed runs, the five-split refit and the learning curve. Five-run
refits use a longer schedule (10 epochs, rate $10^{-5}$) and cross-fitting a
shorter one (Appendix~\ref{sec:s_analyses}). The two large encoders use batch
size 8 in training and validation.}
\label{tab:training_setup}
\end{table}

\paragraph{Fine-tuning and software.} We fine-tune all parameters with the AdamW optimiser
\citep{loshchilov2019decoupled} and, over the first token, a two-output
classification head whose softmax probability of a positive outcome is the
predicted risk, with no recalibration. Of the $7{,}226$ essays, $3.3\%$ to $4.3\%$
exceed 512 tokens, depending on the tokeniser. BioBERT and ClinicalBERT ship
no tokeniser configuration, so we load them cased, as they were pretrained. We
use seed 42 as the primary seed and for the test split, 43 for the validation
split, 1 to 4 for further runs and 1 to 5 only for the five-split refit's test
splits. We pin \texttt{scikit-learn} 1.9.0 \citep{pedregosa2011scikit},
\texttt{PyTorch} 2.3.1 \citep{paszke2019pytorch}, \texttt{transformers} 4.45.2
\citep{wolf2020transformers}, \texttt{MLstatkit} 0.1.91 \citep[DeLong tests,][]{huang2025mlstatkit}, \texttt{Optuna} 4.9.0
\citep{akiba2019optuna}, \texttt{pycaleva} 0.8.2 \citep[calibration belt,][]{weigl2024pycaleva} and \texttt{statsmodels}
0.14.6 \citep{seabold2010statsmodels}. Each encoder run uses one V100 GPU
(16\,GB for the seven encoders of Table~\ref{tab:results}, 32\,GB for the other
four). Each 7B-to-9B LLM runs on one 32\,GB V100, and Qwen2.5-32B on
four.

\paragraph{Model checkpoints.} Table~\ref{tab:checkpoints} lists every
Hugging Face checkpoint.

\begin{table}[t]
\centering
\footnotesize
\begin{tabular*}{\columnwidth}{@{\extracolsep{\fill}}l@{}}
\toprule
Hugging Face model \\
\midrule
\multicolumn{1}{@{}l}{Fine-tuned encoders (Table~\ref{tab:results})} \\
\quad \texttt{bert-base-uncased} \\
\quad \texttt{roberta-base} \\
\quad \texttt{emilyalsentzer/Bio\_ClinicalBERT} \\
\quad \texttt{dmis-lab/biobert-base-cased-v1.1} \\
\quad \texttt{mental/mental-bert-base-uncased} \\
\quad \texttt{mental/mental-roberta-base} \\
\quad \texttt{Amalq/stress-roberta-base} \\
\multicolumn{1}{@{}l}{Further encoders (Table~\ref{tab:extended})} \\
\quad \texttt{bert-base-cased} \\
\quad \texttt{microsoft/deberta-v3-base} (184M) \\
\quad \texttt{microsoft/deberta-v3-large} (434M) \\
\quad \texttt{roberta-large} (355M) \\
\multicolumn{1}{@{}l}{Frozen embeddings (with \texttt{bert-base-uncased})} \\
\quad \texttt{BAAI/bge-large-en-v1.5} \\
\multicolumn{1}{@{}l}{Zero-shot LLMs (Figure~\ref{fig:prompt})} \\
\quad \texttt{Qwen/Qwen2.5-7B-Instruct} \\
\quad \texttt{Qwen/Qwen2.5-32B-Instruct} \\
\quad \texttt{google/gemma-2-9b-it} \\
\quad \texttt{meta-llama/Llama-3.1-8B-Instruct} \\
\bottomrule
\end{tabular*}
\caption{Model checkpoints. The seven encoders of Table~\ref{tab:results}
have $108$ to $125$ million parameters.}
\label{tab:checkpoints}
\end{table}

\paragraph{Task-adaptive pretraining.} We further pretrain each adapted
encoder once by masked language modelling, with no outcome labels, on
$9{,}063$ essays: the training and validation rows plus the $3{,}283$ outside
the analytic sample, excluding test essays by participant identifier. We run
twenty epochs at learning rate $5\times10^{-5}$, batch size 8 and mask
probability $0.15$, monitor the loss on $500$ held-out essays and keep the
last epoch. DeBERTa-v3 \citep{he2023debertav3} was pretrained by
replaced-token detection, and \texttt{transformers} 4.45.2 does not load
weights for its masked-language-model head, which therefore starts from
random weights. DeBERTa-v3's stock and adapted AUCs are thus less comparable
than the other encoders'. The adapted encoders are fine-tuned at
the same five seeds and on the same schedule as the stock ones but have seen
more text, including the validation essays that select their best epoch.

\paragraph{TF-IDF and frozen embeddings.} We fit TF-IDF on the training
rows only, with word 1--2-grams, $20{,}000$ features, minimum document
frequency 2, sublinear weighting and English stopwords removed. Tied n-gram
counts at the vocabulary cut-off make its AUC vary across machines by up to
about $0.005$. Frozen BERT uses the first-token (\texttt{[CLS]}) vector of
each essay's first 512 tokens, whereas frozen BGE averages over the same
tokens despite its model card recommending \texttt{[CLS]}.

\paragraph{Zero-shot LLMs.} Figure~\ref{fig:prompt}'s line breaks and columns
are layout only. We cap each essay at its first $1{,}200$ words and decode
greedily (at most $64$ new tokens), otherwise keeping each checkpoint's
shipped generation settings (Qwen2.5: repetition penalty $1.05$). A reply
parses only if all four ratings are on their 1-to-5 scale (parse rates:
Table~\ref{tab:llmsurvey}), and an unparsed essay takes the training-set mean
of the averaged rating. A univariate logistic regression fitted on the
training rows maps the rating to predicted risk.

\subsection{Analysis protocols}
\label{sec:s_analyses}

\paragraph{Intervals, tests and calibration.} Bootstrap replicates resample
positives and negatives separately, 95\% CIs are percentile intervals, and
each $\Delta$AUC compares both models on the same resampled rows. DeLong tests
are two-sided, and AUC-PR is scikit-learn's average precision.
For the sex gap of Section~\ref{sec:diagnostics}, we resample the test rows of
men ($695$) and women ($751$) independently. The
men-minus-women AUC gap is $0.153$ $[0.056, 0.242]$ for the baseline and
$0.136$ $[-0.004, 0.264]$ for MentalBERT on the primary seed. The Spearman correlation of
Section~\ref{sec:outcome_robust} relates the baseline's predicted risk to the
full Malaise score on the test rows. The slope in
Figure~\ref{fig:calibration} comes from a logistic regression of the outcome on
the logit of the predicted risk (a slope above $1$ means predicted risks
spread too little). At $t^*$ the baseline flags $211$ test rows, $37$ of them positive
(recall $0.374$).

\paragraph{Incremental value.} For Section~\ref{sec:incremental}, we fit the
baseline on the training rows with and without one added text score. A trained
model's score enters as the logit of its predicted risk, whereas a zero-shot
LLM's averaged rating enters standardised on the training rows. $\Delta$AUC and DeLong tests
compare the two L2-penalised fits ($C=1$) on the test rows. DeLong tests can be
conservative for nested models fitted and tested on the same rows
\citep{demler2012misuse}, so a likelihood-ratio
test also compares unpenalised refits on the training
rows. Each cross-fitting run, per fold or final, uses the shorter schedule
(four epochs at rate $10^{-5}$) and selects its epoch on a stratified 20\% of its rows. We
cross-fit TF-IDF and the frozen embeddings the same way. The cross-fitted TF-IDF keeps $10{,}000$
features, drops terms in fewer than 5 or over 95\% of essays, and uses no
sublinear weighting or stopword removal. Alone, the three cross-fitted scores
reach AUC $0.624$ (TF-IDF), $0.586$ (frozen BERT) and $0.634$ (frozen BGE).

\begin{table}[t]
\centering
\tabstyle
\begin{tabular*}{\columnwidth}{@{\extracolsep{\fill}}lccccc@{}}
\toprule
Outcome cut & 6 & 7 & 8 & 9 & 10\phantom{$^\dagger$} \\
Test positives & 218 & 154 & 99 & 59 & 37\phantom{$^\dagger$} \\
\midrule
Baseline, relabelled & 0.690 & 0.712 & 0.737 & 0.733 & 0.739\phantom{$^\dagger$} \\
Baseline, refitted & 0.693 & 0.714 & 0.737 & 0.734 & 0.746\phantom{$^\dagger$} \\
\midrule
\multicolumn{6}{@{}l}{Baseline minus model AUC-ROC, both relabelled} \\
BERT-base & 0.043 & 0.051 & 0.068 & 0.069 & 0.048$^\dagger$ \\
RoBERTa-base & 0.068 & 0.086 & 0.117 & 0.104 & 0.096\phantom{$^\dagger$} \\
ClinicalBERT & 0.052 & 0.064 & 0.070 & 0.064\rlap{$^\dagger$} & 0.033$^\dagger$ \\
BioBERT & 0.051 & 0.062 & 0.066 & 0.059\rlap{$^\dagger$} & 0.029$^\dagger$ \\
MentalBERT & 0.072 & 0.095 & 0.118 & 0.122 & 0.094$^\dagger$ \\
MentalRoBERTa & 0.049 & 0.065 & 0.090 & 0.085 & 0.086$^\dagger$ \\
StressRoBERTa & 0.043 & 0.063 & 0.076 & 0.056\rlap{$^\dagger$} & 0.058$^\dagger$ \\
TF-IDF & 0.073 & 0.080 & 0.092 & 0.088 & 0.094\phantom{$^\dagger$} \\
\bottomrule
\end{tabular*}
\caption{Primary-seed test AUC-ROC by outcome cut for models trained at cut 8
(relabelled) or at the cut (refitted).
$^\dagger$Paired DeLong $p \geq 0.05$ (uncorrected).}
\label{tab:cutsweep}
\end{table}

\paragraph{Learning curve and learning-rate sweep.} The learning curve
(Figure~\ref{fig:learning_curve}) refits the baseline, BERT-base and MentalBERT
on the same stratified training subsamples of six sizes ($578$ to $4{,}624$
rows). Each seed (42 and 1 to 4) draws one subsample for all three models and
initialises the encoder runs, which
select their epoch on all $1{,}156$ validation rows. For MentalBERT, 12
five-epoch Optuna trials search learning rates in $[10^{-6}, 10^{-4}]$ (log
scale, tree-structured Parzen estimator, first ten drawn at random). The best
rate, $5.4\times10^{-5}$, raises validation AUC from $0.588$
(six-epoch recipe, Table~\ref{tab:training_setup}) to $0.631$, against the baseline's $0.708$.

\end{document}